\documentclass[pmlr]{jmlr}

\usepackage{longtable}

\usepackage{booktabs}
\usepackage[load-configurations=version-1]{siunitx} 
\usepackage{mathtools}
\usepackage{bm}
\usepackage{float}  
\usepackage{enumitem}
\usepackage{subcaption}  
\usepackage{multirow}

\newcommand\clearrow{\global\let\rowmac\relax}

\usepackage{xspace}
\newcommand{\mname}
{\texttt{SLEEPER}\xspace}
\theorembodyfont{\upshape}
\theoremheaderfont{\scshape}
\theorempostheader{:}
\theoremsep{\newline}

\jmlrvolume{106}
\jmlryear{2019}
\jmlrworkshop{Machine Learning for Healthcare}

\title[\mname]{\mname: interpretable Sleep staging via Prototypes from Expert Rules}

\author{\Name{Irfan Al-Hussaini} \Email{alhussaini.irfan@gatech.edu} \\
      \addr School of Electrical and Computer Engineering\\
      Georgia Institute of Technology\\
      Atlanta, GA, USA
      \AND
      \Name{Cao Xiao} \Email{cao.xiao@iqvia.com} \\
      \addr Analytics Center of Excellence\\
      IQVIA\\
      Cambridge, MA, USA
      \AND
      \Name{M. Brandon Westover} \Email{mwestover@mgh.harvard.edu} \\
      \addr Department of Neurology\\
      Massachusetts General Hospital\\
      Boston, MA, USA 
      \AND
      \Name{Jimeng Sun} \Email{jsun@cc.gatech.edu} \\
      \addr School of Computational Science and Engineering\\
      Georgia Institute of Technology\\
      Atlanta, GA, USA} 

\begin{document}

\maketitle

\begin{abstract}
Sleep staging is a crucial task for diagnosing sleep disorders. It is tedious and complex as it can take a trained expert several hours to  annotate just one patient's polysomnogram (PSG) from a single night. Although deep learning models have demonstrated state-of-the-art performance in automating sleep staging, interpretability which defines other desiderata, has largely remained unexplored. In this study, 
we propose {\it Sleep staging via Prototypes from Expert Rules} (\mname), which combines deep learning models with expert defined rules using a prototype learning framework to generate simple interpretable models.
In particular, \mname utilizes sleep scoring rules and expert defined features to derive prototypes which are embeddings of PSG data fragments via convolutional neural networks. The final models are simple interpretable models like a shallow decision tree defined over those phenotypes. We evaluated \mname using two PSG datasets collected from sleep studies and demonstrated that \mname could provide accurate sleep stage classification comparable to human experts and deep neural networks with about 85\% ROC-AUC and .7 $\kappa$. 
\end{abstract}

\section{Introduction}
Sleep disorders such as sleep apnea and insomnia affect over 50 to 70 million US adults,  many of whom are undiagnosed ~\citep{sleepstat}. The central diagnostic test is through sleep studies which involve collecting and analyzing  polysomnograms (PSG) data of patients during sleep.
Sleep staging  is the most important task for diagnosing sleep disorders such as insomnia, narcolepsy or sleep apnea \citep{nature}. Typically neurologists will visually inspect
multivariate PSG time series and provide manual scores of sleep stages such as wake, rapid eye movement (REM), non-REM stage N1 to N3.
Such a visual task is cumbersome and requires sleep experts to manually inspect PSG data recorded during the whole sleep study. It can take several hours to annotate one patient's record during a single night. To alleviate this limitation, there has been considerable effort over the years to develop deep learning methods to automate the sleep scoring task due to their promising performances. Recent research include developing artificial visual perception using convolutional neural networks (CNN) \citep{cnn},  recurrent neural networks \citep{rnn}, recurrent convolutional neural networks~\citep{10.1093/jamia/ocy131} and deep belief nets \citep{unsupervised}. 
Although deep learning models can produce accurate sleep staging classification, they are often treated as black-box models that lack interpretability and transparency of their inner working~\citep{DBLP:journals/corr/Lipton16a}. This can limit the adoption of the deep learning models in practice because clinicians often need to understand the reason behind each classification to avoid data noises and unexpected biases.

On the other hand, current clinical practice at sleep labs rely on the American Academy of Sleep Medicine (AASM) sleep scoring manual \citep{berry2012aasm}, which are interpretable for clinical experts but often vague and not computationally precise. Furthermore, the real data are much more heterogeneous and noisy, which lead to more difficult cases to score. As a result, even after certification, technicians often need to acquire multiple years of working experiences in scoring real-patient data at sleep labs before their scores can be trusted.

Can we develop models that are as {\bf interpretable} as the sleep scoring manual but as {\bf accurate} as the black-box neural network models? 
To acquire such a sleep staging model that can produce both accurate and interpretable results, we propose a method based on {\bf prototype learning}, which is an interpretable model inspired by case-based reasoning~\citep{kolodner1992introduction}, where observations are classified based on their proximity to a prototype point in the dataset.  Many machine learning models have incorporated prototype concepts~\citep{priebe2003classification,bien2011prototype,kim2014bayesian}, and  learn to compute prototypes (as actual data points or synthetic points) that can represent a set of similar points. These prototypes provide intuitive understanding of the classifications. Prototype learning also had successes in deep learning models  ~\citep{DBLP:journals/corr/SnellSZ17,DBLP:journals/corr/abs-1710-04806}. The challenges to develop prototype learning methods with deep learning include  
\begin{enumerate}
    \item the resulting models are not necessarily interpretable as the final models are often still complex neural networks;
    \item those models do not capture existing domain knowledge such as scoring rules from the training manual. 
\end{enumerate}

In this work, we propose {\it Sleep staging via Prototypes from Expert Rules} (\mname). 
\mname combines deep learning models with expert defined rules via a prototype learning framework to generate simple interpretable models such as shallow decision trees and logistic regression models.
In particular, \mname utilizes sleep scoring rules and expert defined features to derive prototypes which are embeddings of  polysomnogram (PSG) data fragments via convolutional neural networks. The final models are still simple interpretable models like a shallow decision tree or logistic regression defined over those phenotypes.

\paragraph{Technical Significance}
Although deep learning models have demonstrated state-of-the-art performance in sleep staging, their interpretability has largely remained unexplored. Interpretability helps determine the extent of desiderata beyond performance metrics such as fairness, privacy, reliability, robustness, causality, usability and trust \citep{doshi2017towards}. In the current study, to achieve accurate but much more interpretable sleep stage classification, we develop a framework that first jointly embeds both multivariate PSG data and the staging rules followed by experts into the same latent space using CNN, so that relevance scores between each rule and data prototype can be computed using normalized cosine similarity. 
It then performs staging classification using decision tree and learns staging rules along with relevant prototypes. The results include both expert rules and PSG prototypes, which mimics the visual inspection mechanism of clinical experts.

\paragraph{Clinical Relevance}
Dysfunctional sleep can lead to multiple medical conditions including cardiovascular, metabolic and psychiatric disorders \citep{nature}). Sleep deprivation in the form of insomnia affects 10-15\%  of the adult population causing distress and impairment \citep{insomnia_percentage} with effects ranging from poor memory to increased susceptibility to motor vehicle accidents \citep{accident}. Sleep staging is the most important precursor to sleep disorder diagnosis. However, manual sleep staging is labour intensive and expensive. Computational Sleep Stage Scoring can amortize the cost of diagnosing sleep disorders. Although automatic sleep staging has been explored in depth, interpretation of resulting models remain unexplored. \mname provides a set of clinically meaningful phenotypes, for each prediction. The phenotypes, referred to as prototypes, are derived through rules set forth in \textit{The AASM Manual for the Scoring of Sleep and Associated Events}~\citep{berry2012aasm} and augmented by suggestion from sleep experts. Our resulting shallow decision trees can potentially enhance the training of sleep technicians to learn complex phenotypes related to sleep stages via intuitive explanation. 


\section{\mname Method}

\subsection{Method Overview}

\mname identifies sleep stages on PSG data via interpretable classification models over explainable patterns extracted by expert defined rules. The input data are multi-channel PSG signals segmented into 30-second epochs in the form of multivariate time series data, denoted as 
$\mathcal{X} = \{\bm{X}_1, \cdots, \bm{X}_N\}$, where each epoch $\bm{X}_n \in \mathbb{R}^{9\times6,000}$ is 30 seconds long and contains 9 physiological signals recorded at a frequency of 200Hz. Each epoch $\bm{X}_n$ has a sleep stage label $y_n \in \{\text{Wake, REM, N1, N2, N3}\}$. Our task is to predict the sequence of sleep stages $\bm{S}=\{s_1, \; \dots \; s_N\}$ based on $\mathcal{X}$ so that they are close to the human labels $\bm{Y}=\{y_1, \; \dots \; y_N\}$. We also aim at providing explainable predictions using interpretable classifiers, which are enhanced with neural networks and expert defined rules.

\begin{figure}[htbp]
  \centering 
  \includegraphics[width=\textwidth]{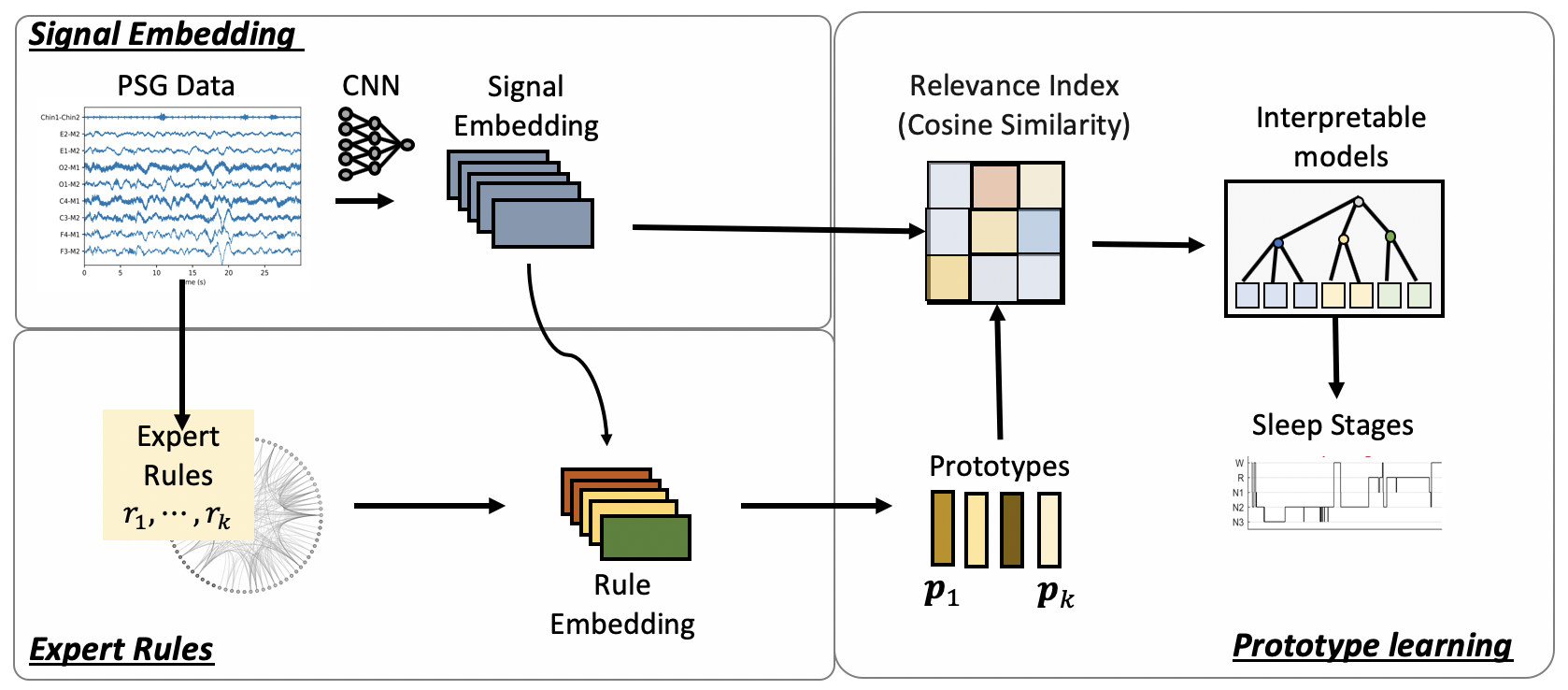} 
  \caption{The \mname framework.}
  \label{fig:flow} 
\end{figure}

As shown in Figure \ref{fig:flow}, \mname comprises of several modules:
\begin{itemize}
    \item {\bf Signal embedding module:} We begin with training the CNN on the end-to-end task of predicting sleep stages using raw PSG data. Afterwards, we remove the last fully-connected layer of the trained CNN and obtain a latent representation, $\bm{h}(\bm{X}_n)$ for epoch $n$.
    \item {\bf Expert rule module:} Concurrently, we use a set of expert rules to encode each epoch into a multi-hot vector, $\bm{R}(\bm{X}_n)=[r_1(\bm{X}_n), \dots r_k(\bm{X}_n)]$, where $k$ is the number of rules and element $r_j(\bm{X}_n)=1 \Leftrightarrow r_j\text{ is satisfied by }\bm{X}_n$. 
    \item {\bf Prototype learning module:} The input encoded by rules and CNN embeddings are combined to form prototypes, $\bm{P} = \{ \bm{p}_1, \dots \, \bm{p}_k \}$, defining each rule in the high-dimensional space of CNN embeddings. Next, the prototypes are used to generate a normalized similarity index for each epoch, $\bm{X}_n$, with each rule, $r_j$. These similarity indices are used to train an interpretable classifier such as decision trees or logistic regression. 
\end{itemize}

\subsection{Dataset}
To evaluate the performance of \mname, we conducted experiments using two datasets.\\

\noindent\textbf{MGH} This refers to a dataset containing PSG recordings of 2000 subjects from Massachusetts General Hospital. The MGH Institutional Review Board approved retrospective analysis of the clinically acquired data without requiring additional consent. The data was randomly selected from a mixture of diagnostic and split night recordings collected from patients whose ages range from $42$ years old to $64$ years old, with an average age of $53$. \\

\noindent\textbf{ISRUC} This refers to the publicly available ISRUC data \citep{isruc}. The ISRUC dataset contains PSG recordings of 100 subjects with evidence of having sleep disorders from Sleep Medicine Centre of the Hospital of Coimbra University (CHUC). The data was collected from 55 male and 45 female subjects, whose ages range from $20$ years old to $85$ years old, with an average age of $51$.\\

The recordings from both dataset were segmented into epochs of 30 seconds and visually scored by sleep technologists according to the guidelines of AASM \citep{berry2012aasm}. The PSG of both datasets include six EEG channels (F3, F4, C3, C4, O1 and O2), two Electrooculography (EOG) channels (E1 and E2) and a single Electromyography (EMG) channel, each referenced to the contralateral mastoid referred to as M1 and M2, or A1 and A2. Additionally, the ISRUC dataset includes scores by \textit{two} sleep technologists. We can thus compare the agreement level between two experts and that between an expert and our algorithms.



\subsection{Expert Rule Embedding} \label{sec:feature}
Majority of the rules in the guideline for sleep technicians \citep{berry2012aasm} are vague. For example, LAMF, Low Amplitude Mixed Frequency are shown through multiple visual examples representing samples of the time domain signal but does not specify a threshold for low amplitude or the power distribution across different frequencies. 
As a result, it is not possible to computationally implement those rules with certainty. 

Our approach alleviates the need for discrete boundaries by creating clusters based on the density of the features in each epoch. We incorporate expert suggestion to supplement the technical guidelines in the AASM manual \citep{berry2012aasm}. Using this rule augmentation procedure, a set of $240$ rules, $\bm{R}'=\{r'_1, \; \dots \; r'_{240}\}$, are defined. Note that those rules are not directly associated with sleep stage labels like the AASM training manual. Instead, the rules define meaningful phenotypes and the similarity with these phenotypes are used as input features to train sleep stage classifiers later, which lead to robust predictions.

The underlying features in those rules are described below, along with the channels utilized for each feature and the corresponding clustering scheme:\\

\noindent\textbf{Sleep spindles} are bursts of oscillatory signals originating from the thalamus and depicted in EEG \citep{spindle}. It is a discriminatory feature of N2. We used the method proposed by \citep{spindle_algo} and \citep{yasa} to extract spindles from contralateral signal pairs resulting in: 
(1) Number of sub-rules: 3 channel pairs with 4 groups for each channel; (2) Channel pairs: i. F3 \& F4, ii. C3 \& C4, iii. O1 \& O2, and (3) Groups: i. $>3$s, ii. $>6$s, iii. $>12$s, iv. $>18$s in an epoch. In total, we have $3 \times 4 = 12$ binary features for spindles. For example, if both F3 and F4 channels exhibit greater than 12 seconds of spindles, the corresponding group will have a feature value 1, and 0 otherwise. 
\\

\noindent\textbf{Slow wave sleep} (SWS) are distinguished by low-frequency and high-amplitude delta activity. Slow waves are the defining characteristics of N3. We utilize the method proposed by \citep{sws_algo1}, \citep{sws_algo2}, and \citep{yasa} to extract SWS from contralateral signal pairs, including (1) Number of sub-rules: 3 channel pairs with 4 groups for each channel, (2) Channel pairs: i. F3 \& F4, ii. C3 \& C4, iii. O1 \& O2, and (3) Groups: i.  $>3$s, ii. $>6$s, iii. $>12$s, iv. $>18$s in an epoch.\\

\noindent\textbf{Delta, Theta, Alpha, and Beta} are the frequency bands which play differing roles in sleep staging. Delta (0.5-4Hz) waves delineate N3, Theta (4-8Hz) features in N1, Alpha (8-12Hz) and Beta ($>$12Hz) discriminates between Wake and N1. The four bands in EMG determine the muscle tone used to distinguish between REM and Wake. We find the Power Spectral Density (PSD) using multitaper spectrogram \citep{mne1, mne2} in each frequency band and make groups based on the percentile of PSD in the training dataset. (1) Number of sub-rules in each band: 9 channels with 4 groups for each channel, (2) Channels: i. F3, ii. F4, iii. C3, iv. C4, v. O1, vi. O2, vii. E1, viii. E2, ix. Chin EMG, and (3) Groups: i. $<20$th percentile, ii. $<40$th percentile, iii. $<60$th percentile, and iv. $<80$th percentile. In total we have $6\times4$ binary features for each frequency band. For example, if the PSD of F3 across the Alpha band of an epoch is $<20$ th percentile the corresponding group will have feature value 1, otherwise 0.\\

\noindent\textbf{Amplitude} is important in discriminating Wake, REM, N1 and N2. Features used in sleep staging that are marked by distinctive amplitude include K Complexes, Chin EMG amplitude, Low Amplitude Mixed Frequency (LAMF). Since the AASM manual \citep{berry2012aasm} does not declare concrete thresholds, we make groups for each and allow our decision tree to connote significance: (1) Number of sub-rules: 9 channels with 4 groups for each channel, (2) Channels: i. F3, ii. F4, iii. C3, iv. C4, v. O1, vi. O2, vii. E1, viii. E2, ix. Chin EMG, (3) Groups: i. $<20$th percentile, ii. $<40$th percentile, iii. $<60$th percentile, and iv. $<80$th percentile.\\

\noindent\textbf{Kurtosis} denotes the distribution of epochs. Although it is not directly related to any feature used by sleep experts, it helps detect outliers in data such as K Complexes which are rare events marked by a distinctive peak and trough. (1) Number of sub-rules: 9 channels with 4 groups for each channel, (2) Channels: i. F3, ii. F4, iii. C3, iv. C4, v. O1, vi. O2, vii. E1, viii. E2, ix. Chin EMG, (3) Groups: i. $<20$th percentile, ii. $<40$th percentile, iii. $<60$th percentile, and iv. $<80$th percentile.

\paragraph{Phenotype selection} We analyze the efficacy of expert defined rules using ANOVA test and select the most discriminative rules.  This reduces the number of expert rules from $240$ to $96$, where $\bm{R}=\{r_1 , \dots r_{96}\}$ and $\bm{R} \subset \bm{R}'$. The resulting channels, underlying feature and the number of groups in each feature-channel pair are shown in Table \ref{tab:features}.  The results from applying all 96 rules on $N$ epochs lead to a binary {\bf rule assignment matrix} $\bm{R}(\mathcal{X})\in \mathbb{R}^{N\times 96}$, which  forms the basis of the interpretation module of \mname framework. 
\begin{equation}\label{eq:rules}
    \bm{R}(\mathcal{X}) = \begin{pmatrix}
                r_1(\bm{X}_1)    & r_2(\bm{X}_1)  & \dots     \\
                r_1(\bm{X}_2)    & \ddots    &           \\
                \vdots      &           &r_{96}(\bm{X}_N)  
            \end{pmatrix}
\end{equation}
where element $r_j(\bm{X}_i) = 1 \Leftrightarrow$ epoch $\bm{X}_i$ satisfies rule $r_j$, $\bm{X}_i\in \mathcal{X}  $, and $\bm{X}_i \in \mathbb{R}^{9\times6,000}$. These resulting features are further discussed in Section \ref{sec:dim_reduce}.

\subsection{Signal Embedding Generation}
\begin{figure*}
\begin{minipage}[c]{0.49\textwidth}
\centering
    \includegraphics[width=\textwidth, keepaspectratio]{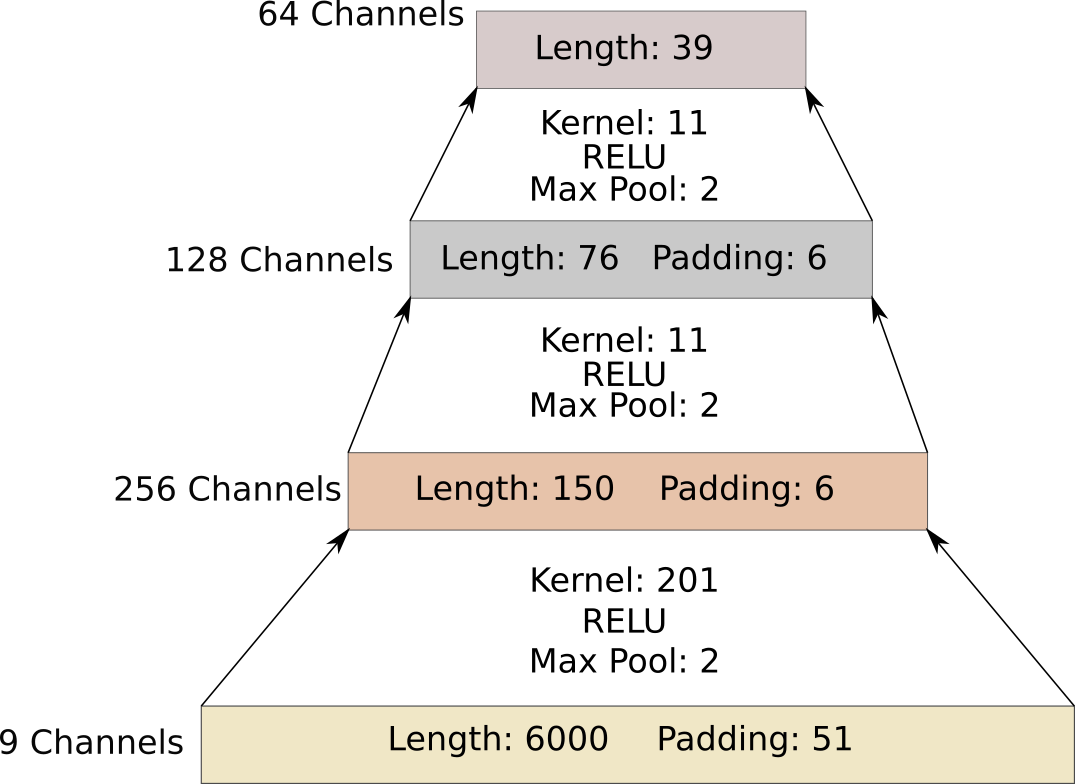}
    \caption{Convolution Layers of CNN }
    \label{fig:cnn}
\end{minipage}
\begin{minipage}[c]{0.49\textwidth}
    \centering
    \includegraphics[ width=\textwidth, keepaspectratio]{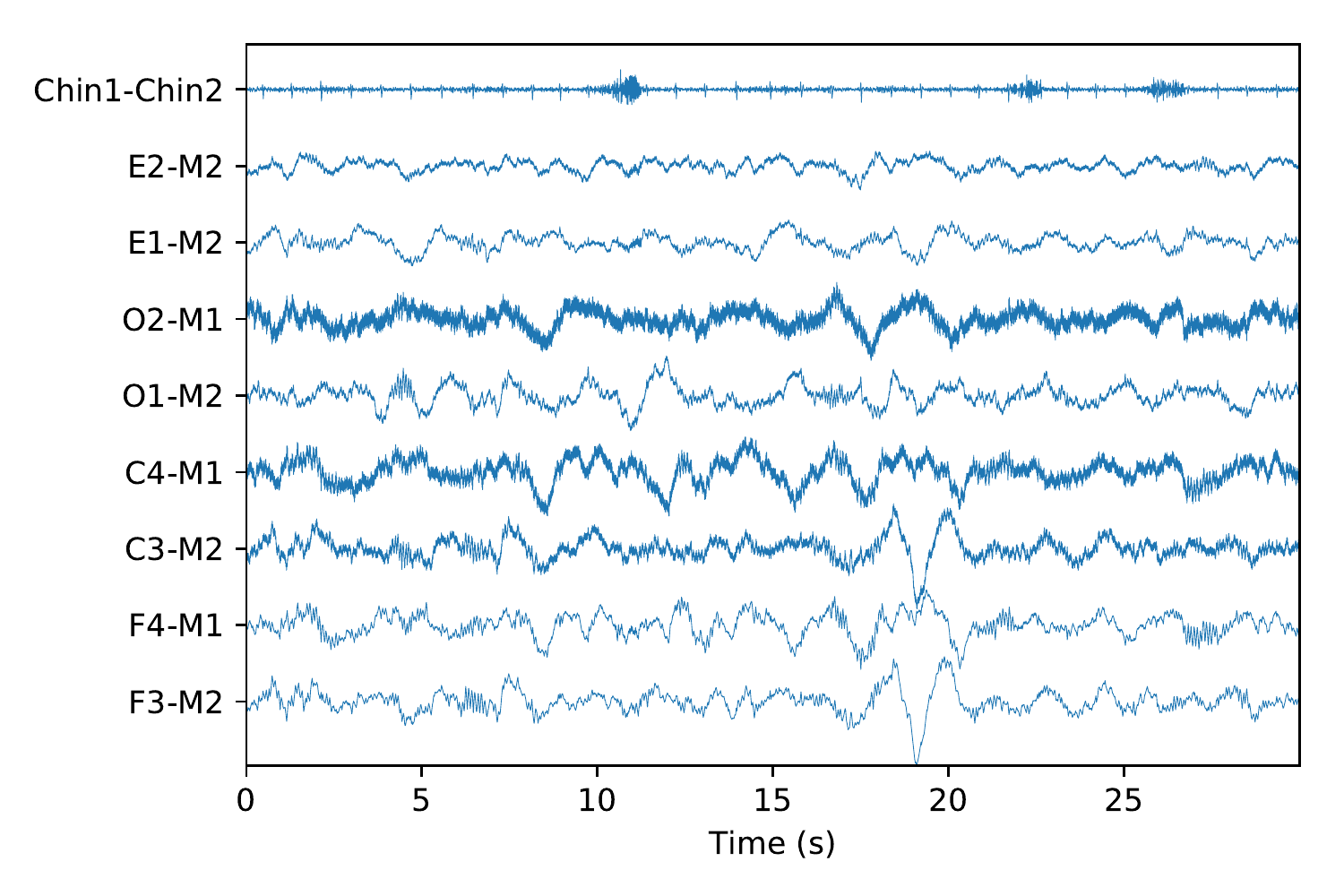}
    \caption{A 30 second epoch having 9 channels}
    \label{fig:epoch}
\end{minipage}
\end{figure*}

The multivariate time series PSG signals were embedded using CNN for capturing translation invariant and complex patterns. The network is composed of 3 convolutional layers as shown in Figure \ref{fig:cnn}. Each convolutional layer is followed by ReLU activation and max pooling. By using a kernel size of 201, the convolutions in the first layer extract features based on 1 second segments of the multivariate time series data.

The output of the final convolutional layer, once flattened, is a vector $\bm{h}(\bm{X}_i) \in \mathbb{R}^{2,496}$. This is followed by a single fully connected layer with softmax activation to predict five different sleep stages:
\begin{eqnarray*} \label{eq:softmax}
    \bm{z}_i &=& \bm{W}^\top \bm{h}(\bm{X}_i) + \bm{b} \\
    \bm{s}_i &=& \text{softmax}(\bm{z}_i)
\end{eqnarray*}
where $\bm{W}\in \mathbb{R}^{2,496\times 5}$ is the weight matrix, $\bm{b}\in\mathbb{R}^5$ is the bias vector, and $\bm{s}_i$ is the estimated probabilities of all 5 sleep stages at epoch $i$. To train the model, we used cross entropy loss in Eq.~\ref{eq:loss}:
\begin{equation} \label{eq:loss}
    L(\bm{y}_i,\; \bm{s}_i) =  - \sum_j^5 \bm{y}_i[j] \; log ( \bm{s}_i[j])
\end{equation} 
where $L(\bm{y}_i,\; \bm{s}_i)$ is the estimated cross entropy loss for epoch $i$ between human labels $\bm{y} \in \mathbb{R}^5$ and the predicted probabilities $\bm{s}\in \mathbb{R}^5$. 
After training on sleep stage prediction, we take the latent representation $\bm{h}(\bm{X})$ of 2,496 dimensions as the PSG signal embedding. 

\subsection{Prototype Learning and Relevance Matching} \label{sec:prototype}
Each of the 96 rules leads to an embedding representation $\bm{p}_j \in \mathbb{R}^{2,496}| j = 1,2,\ldots, 96$. 
Next we describe how to construct the embedding of prototypes using the latent representation of all epochs $\bm{h}(\mathcal{X})$. 
Once we have the rule assignment matrix $\bm{R}(\mathcal{X})$, each prototype representation $\bm{p}_j$ corresponds to the sum of latent embeddings of all the epochs that satisfy the rule $j$. Mathematically, all the prototypes can be computed as 
\begin{equation}\label{eq:prototypes}
    \bm{P} =  \bm{h'}(\mathcal{X})^T \; \bm{R}(\mathcal{X}) 
\end{equation}
where $\bm{P} \in \mathbb{R}^{2,496\times96}$ is all the prototype embeddings, $\bm{h'}(\mathcal{X})\in \mathbb{R}^{N\times2,496}$ the column normalized representation of embedded input $\bm{h}(\mathcal{X})$\footnote{We empirically compared different normalization schemes and this column normalization led to the best performance in our tasks.}.
 Next, we use cosine similarity in the embedding space to rank the similarity of any epoch with rules.
\begin{equation} \label{eq:cosine}
    c_{i, j} = \frac{\bm{h}(\bm{X}_i)^\top \bm{p}_{j} }{  ||\bm{h}(\bm{X}_i)||_2 \; ||\bm{p}_{j}||_2 }
\end{equation}
where $c_{i, j} \in [0,1]$ is the cosine similarity between the $i$th epoch and $j$th phenotype and $\bm{C}(\bm{h}(\mathcal{X}) | \bm{P})\in \mathbb{R}^{N\times96}$ .

 We use these cosine similarity scores  to all prototype embeddings as the input features to simple classifiers. We then train simple and interpretable classifiers such as shallow decision tree and logistic regression to provide the final classifications. When a new PSG, $\bm{X}_{test}$ is given, we find its latent representation using the trained CNN model $\bm{h}(\bm{X}_{test})$, followed by its cosine similarity to existing rule prototypes $\bm{C}(\bm{h}(\bm{X}_{test})|\bm{P})$. Using the simple classifier such as the decision tree, we obtain the predicted sleep stages.
\section{Experiments} 
\subsection{Experiment Setting} 

\noindent\textbf{Baselines.} 
We compare \mname with the following baseline models on both datasets: 
\begin{itemize}
    \item \textbf{Convolutional Neural Network (CNN)}  is the blackbox model used in obtaining the signal embeddings. It serves as the performance upper bound;
    \item \textbf{Rules with Interpretable Classifier} where each epoch is represented by a multi-hot encoded binary rule assignment vector $\bm{R} (\bm{X_i})$ from eqn. \ref{eq:rules} and classification using gradient boosting (GB), decision tree (DT), and logistic regression (LR), respectively.  For the choice of interpretable model in  \mname, we also consider DT, LR and GB.
    \item \textbf{Mimic learning} \citep{mimic} where the soft labels from RCNN are used instead of the original hard labels and a gradient boosting regressor is then trained with those soft labels. 
\end{itemize}

Additionally, on ISRUC dataset we also compare with the following baselines across 5 different sleep stages: (1) Agreement between two sleep experts on the same PSG recordings; (2) Maximum Overlap Discrete Wavelet Transform (MODWT) \citep{isruc, khalighi}; (3) Logistic Smooth Transition Autoregressive (LSTAR) \citep{lstar} (4) Convolutional Neural Network (CNN) \citep{chambon2018deep} (5) RCNN on Spectrogram \citep{10.1093/jamia/ocy131}. Note that (1) and (2) are conducted on the same ISRUC dataset, while (3-5) are on different datasets, which are only for rough comparison.\\
 
\noindent\textbf{Metrics.} We compared testing performance using the following metrics, including accuracy ($Acc$), area under the receiver operator characteristics curve (ROC-AUC), and Cohen's $\kappa$. Here, Cohen's $\kappa$ considers the possibility of assigning the correct sleep stage through random guesses. 
According to \citep{kappa_meaning}, $\kappa > 0.81, \;  0.8 >\kappa > 0.61, \; 0.6 >\kappa >0.41, \; 0.4 >\kappa >0.21, \; 0.2 >\kappa>0.01, \; \kappa <0.01$, means almost perfect, substantial, moderate, fair, slight, less than chance agreement respectively.  We compare the performance across the 5 sleep stages using confusion matrices and class-wise sensitivity ($Sens^{(k)}$), also known as recall. Given expert annotations, $\mathcal{Y}^\prime$ and predicted stages, $\mathcal{Y}$ of size $N$, $k=\{1,2,3,4,5\}$ indicating the sleep stage,
$$Acc = \frac{|\mathcal{Y}\cap \mathcal{Y}'|}{N}, \quad Sens^{(k)} = \frac{ \left| \mathcal{Y}^{(k)} \cap \mathcal{Y}^{\prime (k)} \right|}{\left|\mathcal{Y}^{\prime(k)}\right|} \; $$
$$\kappa = \frac{Acc - p_e}{1-p_e}, \text{ where } p_e = \frac{1}{N^2}\sum_k^5 \left|\mathcal{Y}^{(k)}\right| \left|\mathcal{Y}^{\prime(k)}\right| $$ and $|\mathcal{Y'}^{(k)}|$ ($|\mathcal{Y}^{(k)}|$)is the number of human (algorithm) labels from sleep stage $k$.
 \\

\noindent\textbf{Implementation Details.} We implemented \mname in PyTorch 1.0 \citep{paszke2017automatic} and scikit-learn \citep{scikit-learn}. We train the model using a machine equipped with Intel Xeon e5-2640, 256GB RAM, eight Nvidia Titan-X GPU and CUDA 10.0. While training the CNN, we use batch size of 1 PSG and ADAM as the optimization method.  We train the CNN for 40 epochs. We set the learning rate at $10^{-4}$ and divide the learning rate by 10 once after 10 epochs.

To train the model, we randomly split the data by subjects into training and testing in a 9:1 ratio. For each dataset, we train using the training set to fix model parameters and test on the testing set for performance comparison. To ensure consistent performance across different datasets, we use the same model hyperparameters and underlying feature extraction schema to test both datasets. To evaluate \mname, we consider the following baselines and evaluation metrics.\\
\subsection{Results on Staging Accuracy} 


The experimental results are compared in Table \ref{tab:eval}. On both datasets, \mname performs almost as accurately as the black-box neural network models. Although \mname achieved significant reduction in dimensionality, from $\mathbb{R}^{2,496}$ to $\mathbb{R}^{96}$, the difference in AUC-ROC, accuracy, and Cohen's $\kappa$ to the black-box CNN is relatively small. Moreover, each of those $96$ dimensions are interpretable. \mname-Decision Tree provides a list of normalized indices of length equal to the depth of the tree to indicate similarity with meaningful rules.

\begin{table}[htb]
  \begin{minipage}{\textwidth}
    \centering
\resizebox{\textwidth}{!}{%
\begin{tabular}{l@{\qquad}cc@{\qquad}cc@{\qquad}cc}
  \toprule
  \multirow{2}{*}{\raisebox{-\heavyrulewidth}{Model}} & \multicolumn{2}{c}{Accuracy $(\%)$} & \multicolumn{2}{c}{ROC-AUC $(\%)$} & \multicolumn{2}{c}{Cohen's $\kappa$} \\
  \cmidrule{2-7}
                        &MGH    &ISRUC  &MGH    &ISRUC  &MGH    &ISRUC \\
  \midrule
  \mname -DT     & 78.3 & 78.5 & 85.0 & 84.7 & 0.694 & 0.720 \\
  \mname -LR    & 79.8     & 77.0 & 86.1     & 84.9 & 0.714     & 0.699 \\
  \mname -GBT    & 78.8     & 80.1 & 85.4     & 86.0 & 0.700     & 0.741 \\  
\hline
  Rule \& DT  & 66.1 & 67.1 & 75.7 & 78.2 & 0.510 & 0.564 \\  
  Rule \& LR  & 65.3     & 69.1 &  75.0    & 79.0 & 0.498     & 0.593 \\  
  Rule \& GBT  & 65.8     & 69.3 & 75.3     & 78.8 & 0.508     & 0.594 \\
  Mimic learning - GBT & 67.5     & 62.1 & 78.6     & 76.4 & 0.540     & 0.514 \\
  \hline
  CNN             & 81.6 & 82.4 & 87.4 & 87.8 & 0.742 & 0.772 \\
  \bottomrule
\end{tabular}}
    \caption[LOR]{Model Evaluation.\footnote{96 rules and the corresponding prototypes are used in Rule and \mname respectively} DT: Decision Tree, LR: Logistic Regression, GBT: Gradient Boosting Trees, Rule: Binary Features from Rules, CNN: Convolutional Neural Network}
    \label{tab:eval}
    \end{minipage}
\end{table}

\begin{table}[htb]%
  \begin{minipage}{\textwidth}
\resizebox{\textwidth}{!}{%
\begin{tabular}{l@{\quad}cccccc}
  \toprule
  \multirow{2}{*}{\raisebox{-\heavyrulewidth}{Model}} & \multicolumn{5}{c}{Sensitivity $(\%)$}  \\
  \cmidrule{2-6}
                        &Wake   &REM    &N1     &N2     &N3\\
  \midrule
  \mname -DT\footnote{Depth, $D=9$}             &88.3  &85.3  &26.9  &82.59  &85.6\\
  \mname -LR                  &87.9  &80.6  &27.3  &84.4  &79.1\\
  \mname -GBT                  &88.1  &86.1  &34.5  &83.2  &87.9\\\hline
  Human Expert Agreement     &92.4  &91.2  &55.4  &86.6  &77.4\\
  MODWT \citep{isruc}      &88.3  &81.8  &39.3  &80.2  &83.5\\
  \hline
  CNN \citep{chambon2018deep} &85  &83  &52  &77  &91\\
  LSTAR \citep{lstar}      &88.7  &88.4  &50.3  &85.0  &87.4\\
  RCNN on Spectrogram \citep{10.1093/jamia/ocy131}     &85     &92     &58     &89     &86\\
  \bottomrule
\end{tabular}}
    \caption[for LOF]{Sensitivity across Sleep Stages \footnote{The top 5 rows are results from the same cohort in ISRUC dataset \citep{isruc}}. DT: Decision Tree, GB: Gradient Boosting Trees, LR: Logistic Regression,}
    \label{tab:comp}
    \end{minipage}
\end{table}

\begin{figure*}
    \centering
    \subfigure[Inter-Rater Agreement]
    {
        \centering
        \includegraphics[width=0.45\textwidth]{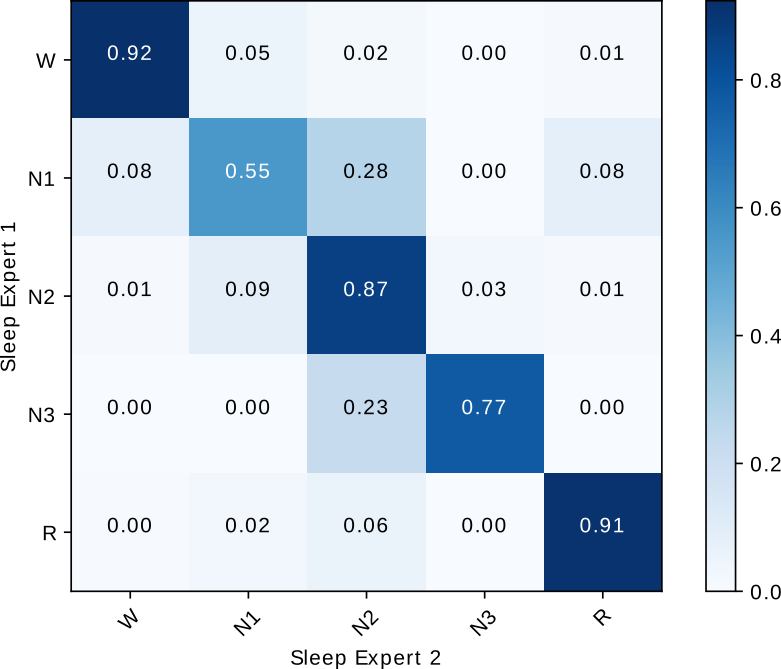}
        \label{fig:inter_rater}
        \hspace{1em}
    }%
    \subfigure[\mname $ $- Decision Tree]
    {
        \hspace{1em}
        \centering
        \includegraphics[width=0.45\textwidth]{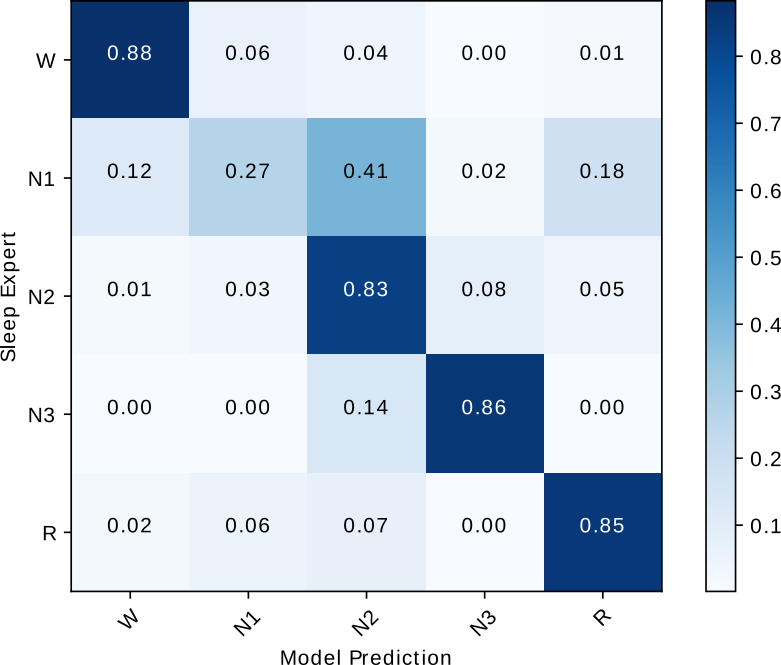}
        \label{fig:conf_prototype}
    }\vspace{1.5em}
    \subfigure[\mname $ $- Gradient Boosting]
    {
        \centering
        \includegraphics[width=0.45\textwidth]{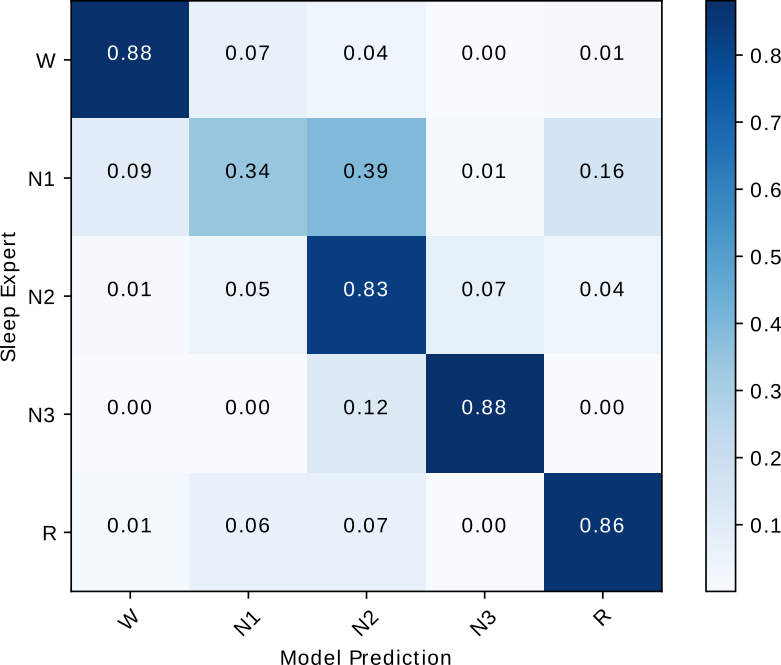}
        \label{fig:conf_gb}
        \hspace{1em}
    }%
    \subfigure[Rules and Decision Tree]
    {
        \hspace{1em}
        \centering
        \includegraphics[width=0.45\textwidth]{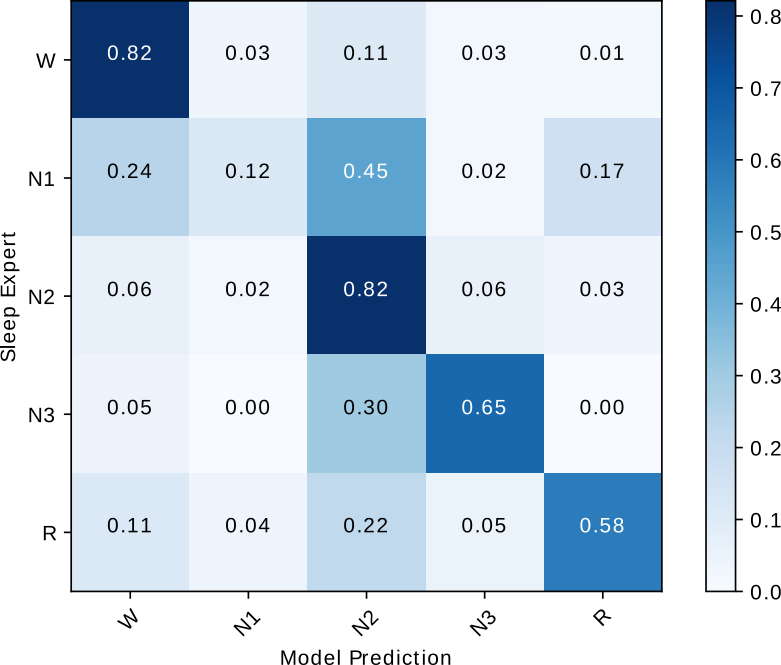}
        \label{fig:conf_tree}
    }\vspace{1.5em}
    \subfigure[CNN]
    {
        \centering
        \includegraphics[width=0.45\textwidth]{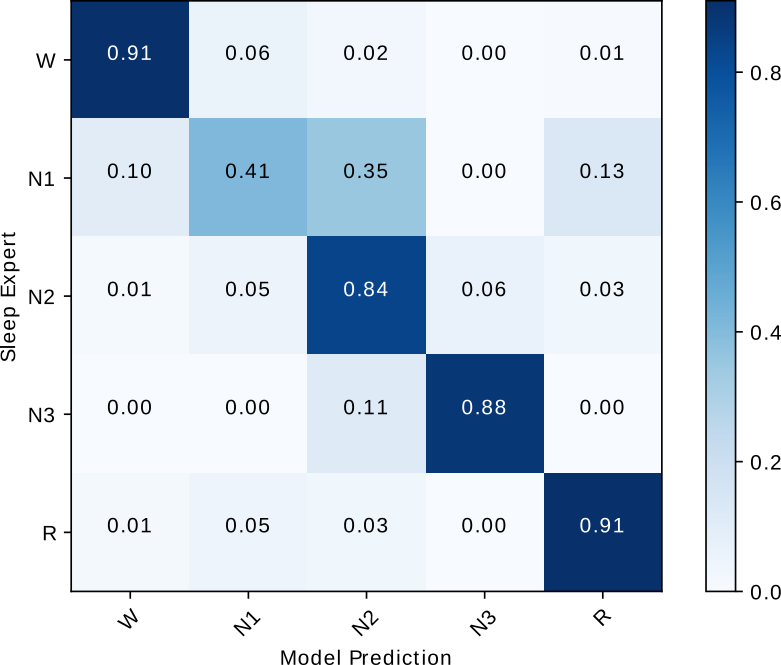}
        \label{fig:conf_1dcnn}
        \hspace{1em}
    }%
    \caption[LOF]{Confusion Matrices on ISRUC dataset}
    \label{fig:conf_mat}
\end{figure*}

The sensitivity in classifying each sleep stage is compared with baselines in Table \ref{tab:comp}. The confusion matrices of our results using ISRUC dataset is shown in Figure \ref{fig:conf_mat}. Agreement between experts are shown in Figure \ref{fig:inter_rater} and \mname using a decision tree in Figure \ref{fig:conf_prototype}.  N1 classification is particularly problematic even for human sleep experts. This is due to significant overlap in underlying criteria with N2. Beyond N2, \mname with Decision Tree surpass the performance of the baseline automatic sleep staging algorithm \citep{khalighi} while also providing interpretation. It exceeds expert agreement by a significant margin for N3. Reasons for this are further discussed in Section \ref{sec:interpretation}.

The agreement between two human experts in assigning sleep stages to our test PSGs in the ISRUC dataset is 83.0\%, with a Cohen $\kappa$ of 0.78. \mname using a decision tree obtains an accuracy of 78.5\%, with a Cohen's $\kappa$ of 0.72 indicating substantial agreement according to the guidelines from \citep{kappa_meaning}.

\begin{figure*}[htb]
    \centering
    \subfigure[ISRUC Dataset]
    {
        \centering
        \includegraphics[width=0.5\textwidth, keepaspectratio]{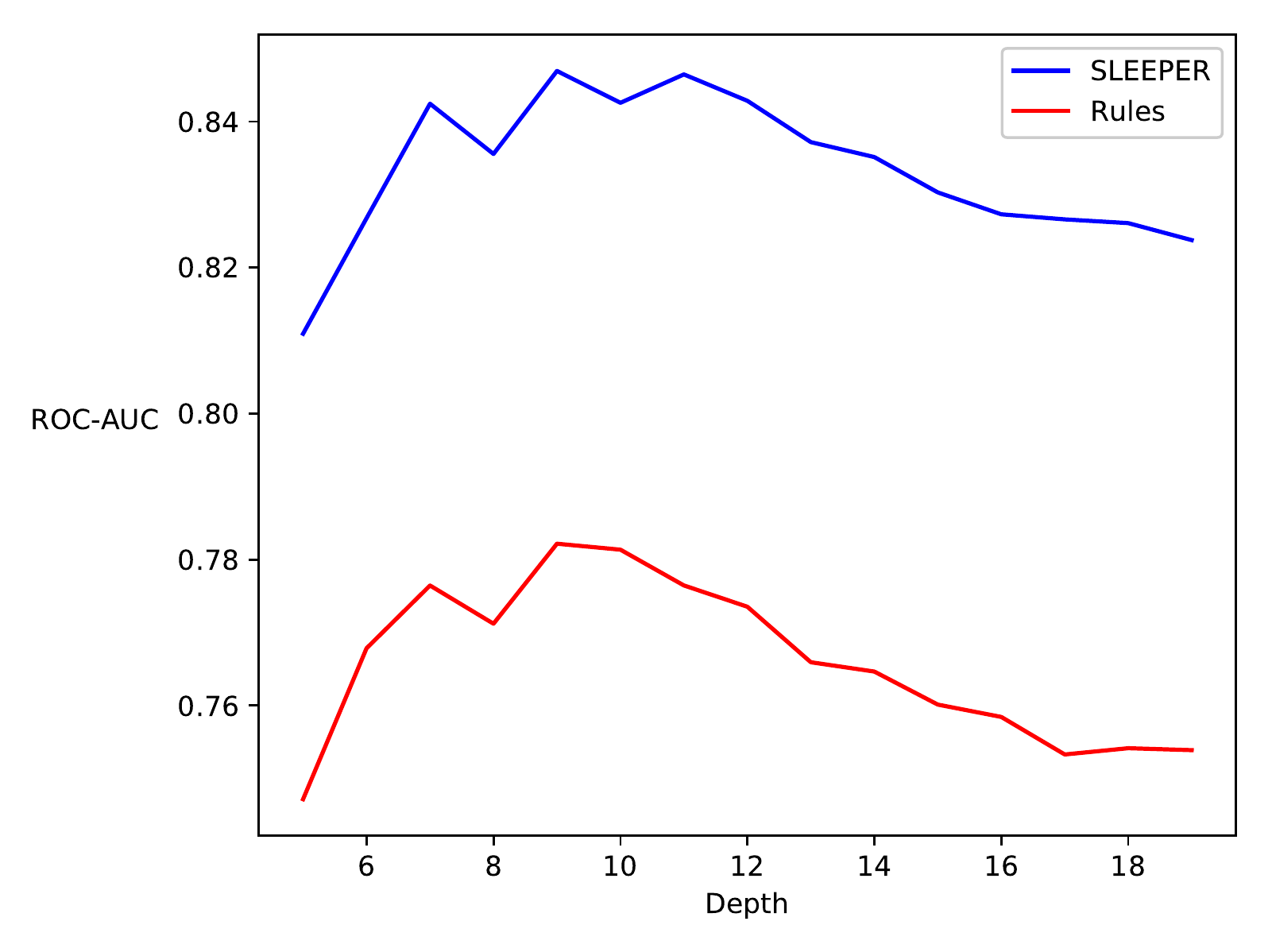}
        \label{fig:tree_depth_isruc}
    }%
    \subfigure[MGH Dataset]
    {
        \centering
        \includegraphics[width=0.5\textwidth]{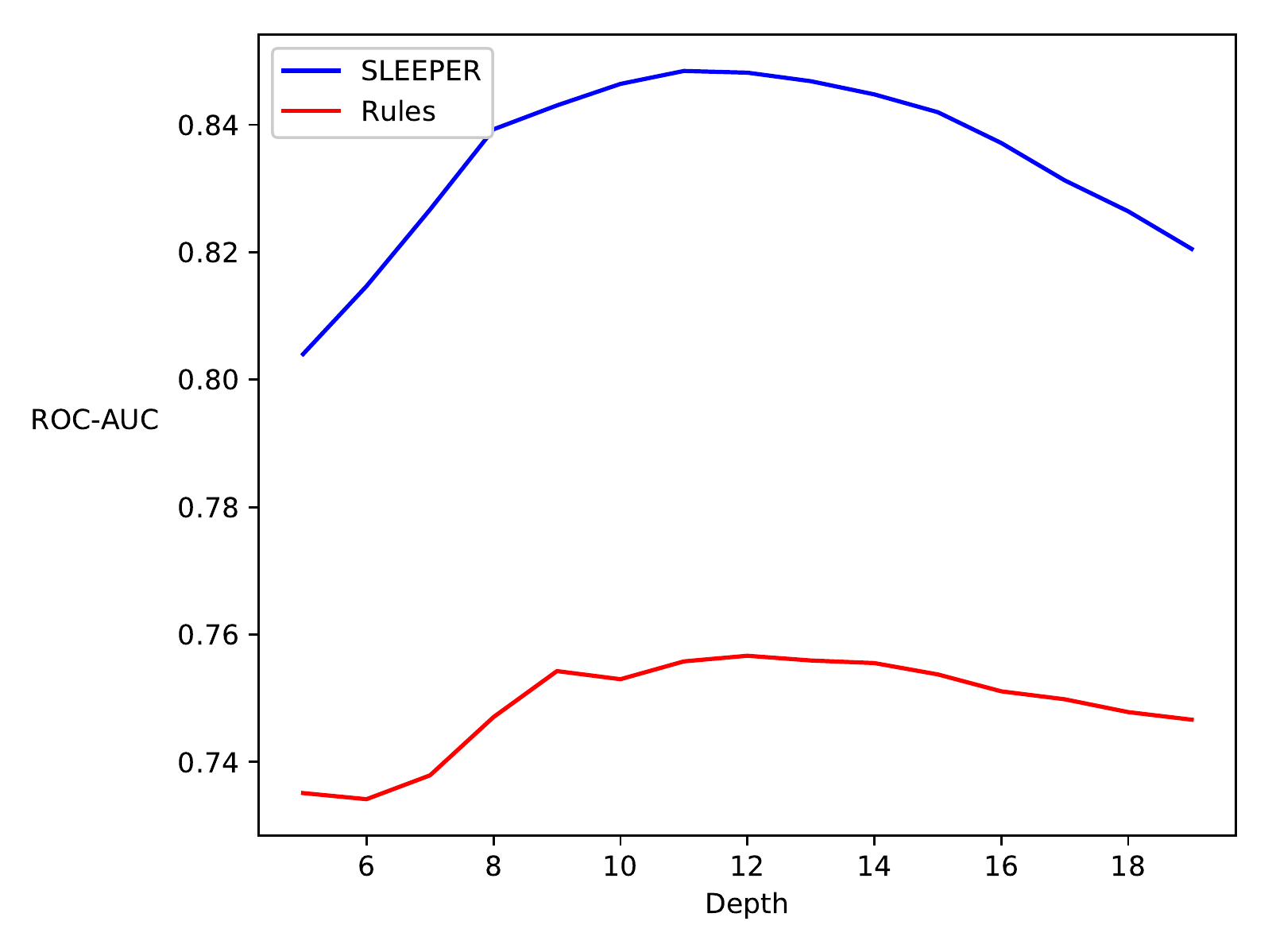}
        \label{fig:tree_depth_isruc_mgh}
    }
\caption{\mname ROC-AUC vs Tree Depth}
\label{fig:tree_deep}
\end{figure*}

Figure \ref{fig:tree_deep} shows the change in ROC-AUC of \mname and rule based method with depth of tree. It shows for trees of depth 9 we obtain ROC-AUC greater than 84\% in both ISRUC and MGH Datasets. This indicates, a group of 9 meaningful prototypes is able to classify the sleep stage in a sample well. Note, the larger size of MGH Dataset results in a much smoother distribution but the overall performance remain similar. This shows robustness across different datasets and the significant performance improvement from rules using \mname.
 
\subsection{Feature Selection Results of Expert Rules}\label{sec:dim_reduce}

\begin{figure*}
    \centering
    \includegraphics[ width=0.49\textwidth, keepaspectratio]{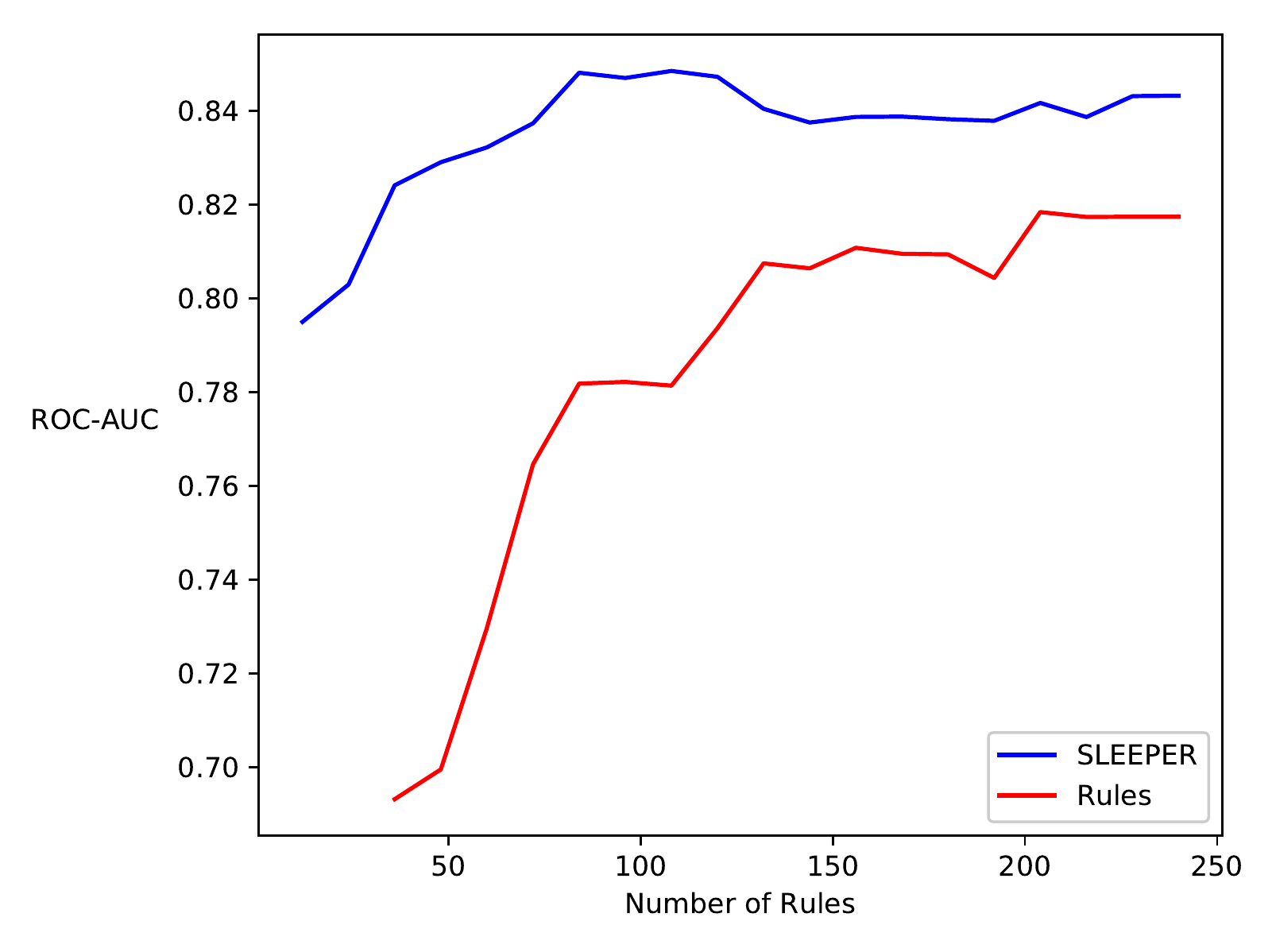}
    \caption{\mname ROC-AUC vs Number of Rules}
    \label{fig:feat_num_roc}
\end{figure*}


\begin{table}
  \begin{minipage}{\textwidth}
\resizebox{\textwidth}{!}{%
\begin{tabular}{l@{\quad}cccccccc}
  \toprule
  \multirow{2}{*}{\raisebox{-\heavyrulewidth}{Channels}} & \multicolumn{8}{c}{Features} \\
  \cmidrule{2-9}
  & Spindle & SWS & Delta & Theta & Alpha & Beta & Kurtosis & Amplitude \\
  \midrule
 F3-A2    &  &3  &4  &  &4  &4  &  &4 \\
 F4-A1    &  &3  &4  &4  &4  &  &  &4\\
 C3-A2    &  &2  &4  &4  &  &4  &  &4\\
 C4-A1    &  &2  &4  &  &  &  &  &4\\
 O1-A2    &  &  &4  &4  &  &  &  &4\\
 O2-A1    &  &  &4  &3  &4  &  &  &4\\
 ROC-A2    &  &  &4  &  &  &  &  &4\\
 LOC-A2    &  &  &  &  &  &  &  &\\
 Chin EMG    &  &  &  &  & &  &  &\\
  \bottomrule
\end{tabular}}
    \caption[LOF]{96 selected rules out of 240 expert rules using ISRUC dataset\footnote{96 selected rules for MGH Dataset is shown in Table \ref{tab:features_mgh} in Supplemental material.}} 
    \label{tab:features}
\end{minipage}
\end{table}
Instead of handpicking due to ambiguity in their significance, we use analysis of variance (ANOVA) models to rank features by significance and reduce the number of rules from 240 to 96. The change in ROC-AUC with number of rules is shown in Figure \ref{fig:feat_num_roc}. It reveals the discrepancy in performance between \mname and the rule based method. 
The selected number of expert rules from each channel-feature pair is shown in Table \ref{tab:features}.
Note that the selected features do not include any features from Spindles and Kurtosis. In particular, sleep spindles have frequency range of 12-14 Hz with a duration of 0.5-1.5 seconds but their use in detecting N2 is practically difficult. This could be due to hidden spindles in other stages \citep{spindle_algo}. And Kurtosis is a common statistical measure but is not directly related to key features described in sleep scoring manual. We also observe the removal of rules using EMG and second EOG channels. EMG recordings are particularly noisy with low amplitude, as shown in the top channel of Figure \ref{fig:epoch}. 

K Complex, another underlying feature in N2 and REM detection, contains a distinctive rise and fall which is larger than the amplitude of regular signal oscillations. K Complexes, unfortunately, are not relibable enough for our use case with an inter-rater $\kappa$ of .51 \citep{kcomplexkappa}. On the other hand, amplitude based prototypes play a big role in \mname. Low Amplitude Mixed Frequency (LAMF) is a feature used for discriminating Wake, N1, N2 and REM. The channels used in detecting LAMF are not mentioned in the guidelines \citep{berry2012aasm}.


\subsection{Interpretation} \label{sec:interpretation}

\begin{figure*}[htb]
 \begin{minipage}{\textwidth}
    \centering
    \includegraphics[ width=\textwidth, keepaspectratio]{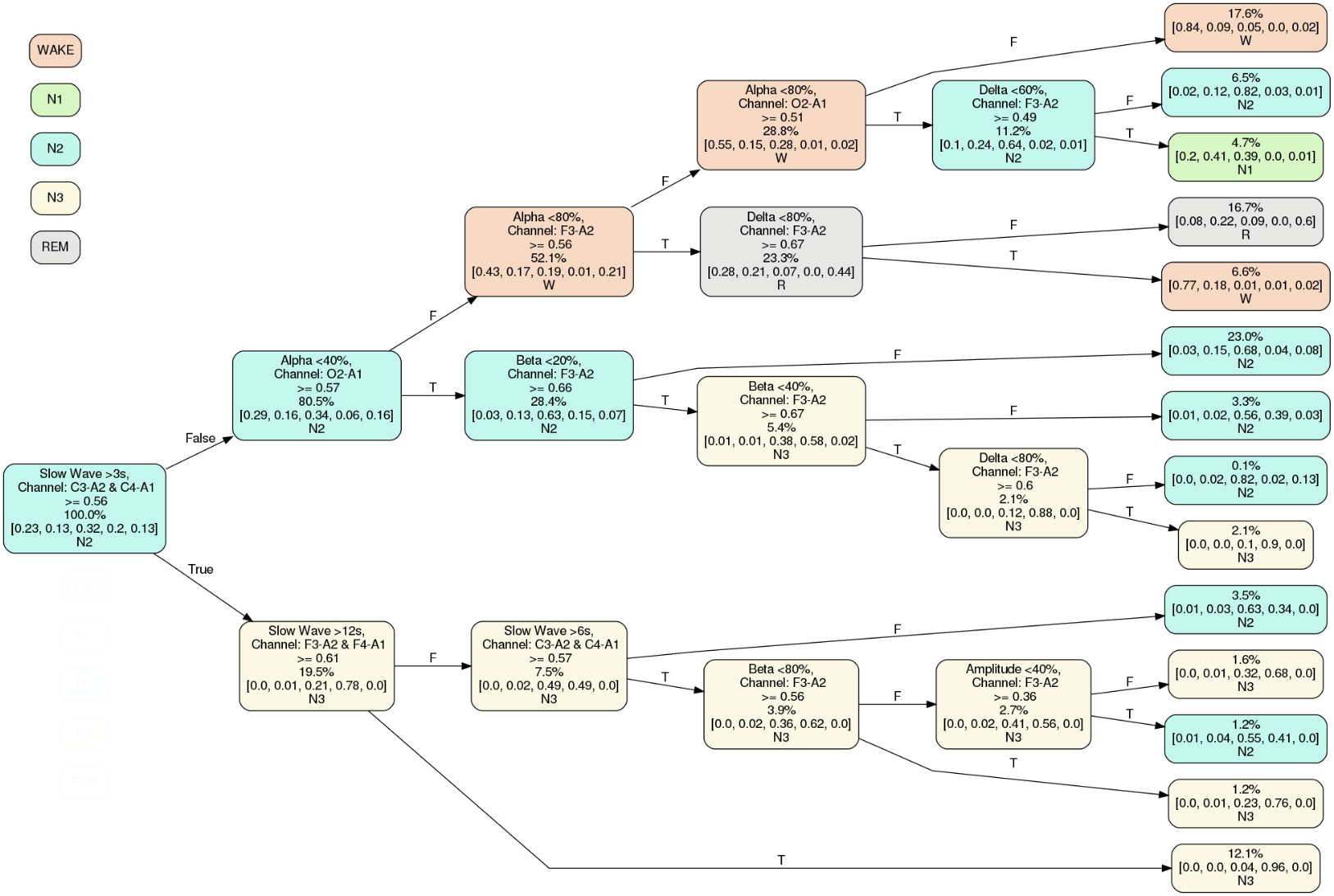}
    \caption[tmp LOF]{\mname-Decision Tree trained from the ISRUC dataset {\let\thefootnote\relax\footnotetext{For example, the root node considers the prototype defining slow waves greater than 3 seconds (row 1) on both C3-A2 and C4-A1 channels (row 2). The cosine similarity threshold is 0.56 (row 3). 100\% (row 4) of all training examples go through this node. [.23, .13, .32, .2, .13] (row 5) are the probabilities of the sleep stages in the training examples in the order of [Wake, N1, N2, N3, R] with N2 (row 6) as majority.}}} 
    \label{fig:tree}
\end{minipage}
\end{figure*}
Figure \ref{fig:tree} shows a decision tree of depth, $D=5$, based on \mname. This obtains an AUC-ROC of 81\%. The leftmost node denotes the root of the tree and the colour indicates the most frequent sleep stage for training data passing through that node. The six rows of each node contains the following: (1) the underlying feature and grouping criteria, (2) the channels used to extract the feature, (3) the cosine similarity with the resulting rule, (4) the percentage of data passing through the node, (5) the ratio of the each sleep stage in data passing through the node in the following order: [Wake, N1, N2, N3, REM], (6) the most frequent sleep stage at the node, in other words, if classification is performed at that node we will assign this label. The leaves on the right contain the same contents as lowest 3 rows at other nodes. 

Analyzing the resulting decision tree reveals some promising aspects of \mname. According to the sleep staging guidelines for human annotators \citep{berry2012aasm}, N3 is distinguished by occurrence of slow waves. One of the underlying features of our rules is slow waves. We created 4 binary features based on the duration of slow wave in each 30s epoch, $>3$s, $>6$s, $>12$s, and $>18$s. The first node creates a split based on cosine similarity $\ge 0.56$ with the prototype, Slow Waves of duration greater than 3s in the Central Channels. Since slow waves are predominant in N3, $78\%$ of training data that satisfied the aforementioned criteria in the next node contains N3, while only $6\%$ of the other child contains N3. The next node restricts the threshold to 12s in the Frontal region. $96\%$ of the resulting leaf node classifying $12.1\%$ of the training dataset was labelled, in agreement with \mname, as N3 by experts.

Furthermore, analyzing the leaves, we observe stages with similar characteristics occur in pairs, like REM and Wake, N3 and N2, N1 and N2. We notice that top right leaf containing Wake is distinguished by Alpha activity in the Occipital Region. This criteria for detecting Wake is mentioned in the guidelines for human annotators \citep{berry2012aasm}.





\section{Conclusion}
Interpretability and accuracy are often trade-off to each other in machine learning modeling. Especially in the age of deep learning, many accurate models are black-box models that do not provide any insights into the reasoning behind predictions. On the other hand, simple models like decision trees often result in inaccurate predictors. In this paper, we present \mname that introduces a deep prototype learning method that provides accurate predictions as well as very simple and intuitive prediction models with a shallow decision tree. We develop and evaluate the methods in the context of sleep staging applications on PSG data from sleep labs. \mname achieves high accuracy in sleep staging tasks comparable to state of the art baselines. A qualitative case study illustrated a simple and intuitive decision tree that can perform accurate sleep staging classification while providing explanation through interpretable rules. 


\section*{Acknowledgments}
This work was supported by the National Science Foundation award IIS-1418511, CCF-1533768 and IIS-1838042, the National Institute of Health award 1R01MD011682-01 and R56HL138415.

\bibliography{mlhc}

\begin{thebibliography}{33}
\providecommand{\natexlab}[1]{#1}
\providecommand{\url}[1]{\texttt{#1}}
\expandafter\ifx\csname urlstyle\endcsname\relax
  \providecommand{\doi}[1]{doi: #1}\else
  \providecommand{\doi}{doi: \begingroup \urlstyle{rm}\Url}\fi

\bibitem[ASA(2019)]{sleepstat}
ASA.
\newblock Sleep statistics - research \& treatments | american sleep assoc.
\newblock \url{https://www.sleepassociation.org/about-sleep/sleep-statistics/},
  2019.

\bibitem[Berry et~al.(2012)Berry, Budhiraja, Gottlieb, Gozal, Iber, Kapur,
  Marcus, Mehra, Parthasarathy, Quan, et~al.]{berry2012aasm}
Richard~B Berry, Rohit Budhiraja, Daniel~J Gottlieb, David Gozal, Conrad Iber,
  Vishesh~K Kapur, Carole~L Marcus, Reena Mehra, Sairam Parthasarathy, Stuart~F
  Quan, et~al.
\newblock Rules for scoring respiratory events in sleep: update of the 2007
  aasm manual for the scoring of sleep and associated events.
\newblock \emph{Journal of clinical sleep medicine}, 8\penalty0 (05):\penalty0
  597--619, 2012.

\bibitem[Bien and Tibshirani(2011)]{bien2011prototype}
Jacob Bien and Robert Tibshirani.
\newblock Prototype selection for interpretable classification.
\newblock \emph{The Annals of Applied Statistics}, pages 2403--2424, 2011.

\bibitem[Biswal et~al.(2018)Biswal, Sun, Sun, Westover, Goparaju, and
  Bianchi]{10.1093/jamia/ocy131}
Siddharth Biswal, Jimeng Sun, Haoqi Sun, M~Brandon Westover, Balaji Goparaju,
  and Matt~T Bianchi.
\newblock {Expert-level sleep scoring with deep neural networks}.
\newblock \emph{Journal of the American Medical Informatics Association},
  25\penalty0 (12):\penalty0 1643--1650, 11 2018.
\newblock ISSN 1527-974X.
\newblock \doi{10.1093/jamia/ocy131}.
\newblock URL \url{https://doi.org/10.1093/jamia/ocy131}.

\bibitem[Carrier et~al.(2011)Carrier, Viens, Poirier, Robillard, Lafortune,
  Vandewalle, Martin, Barakat, Paquet, and Filipini]{sws_algo1}
Julie Carrier, Isabelle Viens, Ga{\'e}tan Poirier, R{\'e}becca Robillard,
  Marjolaine Lafortune, Gilles Vandewalle, Nicolas Martin, Marc Barakat, Jean
  Paquet, and Daniel Filipini.
\newblock Sleep slow wave changes during the middle years of life.
\newblock \emph{European Journal of Neuroscience}, 33\penalty0 (4):\penalty0
  758--766, 2011.

\bibitem[Chambon et~al.(2018)Chambon, Galtier, Arnal, Wainrib, and
  Gramfort]{chambon2018deep}
Stanislas Chambon, Mathieu~N Galtier, Pierrick~J Arnal, Gilles Wainrib, and
  Alexandre Gramfort.
\newblock A deep learning architecture for temporal sleep stage classification
  using multivariate and multimodal time series.
\newblock \emph{IEEE Transactions on Neural Systems and Rehabilitation
  Engineering}, 26\penalty0 (4):\penalty0 758--769, 2018.

\bibitem[Che et~al.(2015)Che, Purushotham, Khemani, and Liu]{mimic}
Zhengping Che, Sanjay Purushotham, Robinder Khemani, and Yan Liu.
\newblock Distilling knowledge from deep networks with applications to
  healthcare domain.
\newblock \emph{arXiv preprint arXiv:1512.03542}, 2015.

\bibitem[De~Gennaro and Ferrara(2003)]{spindle}
Luigi De~Gennaro and Michele Ferrara.
\newblock Sleep spindles: an overview.
\newblock \emph{Sleep medicine reviews}, 7\penalty0 (5):\penalty0 423--440,
  2003.

\bibitem[Dong et~al.(2018)Dong, Supratak, Pan, Wu, Matthews, and Guo]{rnn}
Hao Dong, Akara Supratak, Wei Pan, Chao Wu, Paul~M Matthews, and Yike Guo.
\newblock Mixed neural network approach for temporal sleep stage
  classification.
\newblock \emph{IEEE Transactions on Neural Systems and Rehabilitation
  Engineering}, 26\penalty0 (2):\penalty0 324--333, 2018.

\bibitem[Doshi-Velez and Kim(2017)]{doshi2017towards}
Finale Doshi-Velez and Been Kim.
\newblock Towards a rigorous science of interpretable machine learning.
\newblock \emph{arXiv preprint arXiv:1702.08608}, 2017.

\bibitem[Ghasemzadeh et~al.(2019)Ghasemzadeh, Kalbkhani, Sartipi, and
  Shayesteh]{lstar}
Peyman Ghasemzadeh, Hashem Kalbkhani, Shadi Sartipi, and Mahrokh~G Shayesteh.
\newblock Classification of sleep stages based on lstar model.
\newblock \emph{Applied Soft Computing}, 75:\penalty0 523--536, 2019.

\bibitem[Gramfort et~al.(2013)Gramfort, Luessi, Larson, Engemann, Strohmeier,
  Brodbeck, Goj, Jas, Brooks, Parkkonen, et~al.]{mne2}
Alexandre Gramfort, Martin Luessi, Eric Larson, Denis~A Engemann, Daniel
  Strohmeier, Christian Brodbeck, Roman Goj, Mainak Jas, Teon Brooks, Lauri
  Parkkonen, et~al.
\newblock Meg and eeg data analysis with mne-python.
\newblock \emph{Frontiers in neuroscience}, 7:\penalty0 267, 2013.

\bibitem[Gramfort et~al.(2014)Gramfort, Luessi, Larson, Engemann, Strohmeier,
  Brodbeck, Parkkonen, and H{\"a}m{\"a}l{\"a}inen]{mne1}
Alexandre Gramfort, Martin Luessi, Eric Larson, Denis~A Engemann, Daniel
  Strohmeier, Christian Brodbeck, Lauri Parkkonen, and Matti~S
  H{\"a}m{\"a}l{\"a}inen.
\newblock Mne software for processing meg and eeg data.
\newblock \emph{Neuroimage}, 86:\penalty0 446--460, 2014.

\bibitem[Khalighi et~al.(2013)Khalighi, Sousa, Pires, and Nunes]{khalighi}
Sirvan Khalighi, Teresa Sousa, Gabriel Pires, and Urbano Nunes.
\newblock Automatic sleep staging: A computer assisted approach for optimal
  combination of features and polysomnographic channels.
\newblock \emph{Expert Systems with Applications}, 40\penalty0 (17):\penalty0
  7046--7059, 2013.

\bibitem[Khalighi et~al.(2016)Khalighi, Sousa, Santos, and Nunes]{isruc}
Sirvan Khalighi, Teresa Sousa, Jos{\'e}~Moutinho Santos, and Urbano Nunes.
\newblock Isruc-sleep: a comprehensive public dataset for sleep researchers.
\newblock \emph{Computer methods and programs in biomedicine}, 124:\penalty0
  180--192, 2016.

\bibitem[Kim et~al.(2014)Kim, Rudin, and Shah]{kim2014bayesian}
Been Kim, Cynthia Rudin, and Julie~A Shah.
\newblock The bayesian case model: A generative approach for case-based
  reasoning and prototype classification.
\newblock In \emph{Advances in Neural Information Processing Systems}, pages
  1952--1960, 2014.

\bibitem[Kolodner(1992)]{kolodner1992introduction}
Janet~L Kolodner.
\newblock An introduction to case-based reasoning.
\newblock \emph{Artificial intelligence review}, 6\penalty0 (1):\penalty0
  3--34, 1992.

\bibitem[Krieger(2017)]{accident}
Ana~C Krieger.
\newblock \emph{Social and Economic Dimensions of Sleep Disorders, An Issue of
  Sleep Medicine Clinics, E-Book}, volume~12.
\newblock Elsevier Health Sciences, 2017.

\bibitem[Lacourse et~al.(2019)Lacourse, Delfrate, Beaudry, Peppard, and
  Warby]{spindle_algo}
Karine Lacourse, Jacques Delfrate, Julien Beaudry, Paul Peppard, and Simon~C
  Warby.
\newblock A sleep spindle detection algorithm that emulates human expert
  spindle scoring.
\newblock \emph{Journal of neuroscience methods}, 316:\penalty0 3--11, 2019.

\bibitem[Lajnef et~al.(2015)Lajnef, Chaibi, Eichenlaub, Ruby, Aguera, Samet,
  Kachouri, and Jerbi]{kcomplexkappa}
Tarek Lajnef, Sahbi Chaibi, Jean-Baptiste Eichenlaub, Perrine~Marie Ruby,
  Pierre-Emmanuel Aguera, Mounir Samet, Abdennaceur Kachouri, and Karim Jerbi.
\newblock Sleep spindle and k-complex detection using tunable q-factor wavelet
  transform and morphological component analysis.
\newblock \emph{Frontiers in human neuroscience}, 9:\penalty0 414, 2015.

\bibitem[L\"{a}ngkvist et~al.(2012)L\"{a}ngkvist, Karlsson, and
  Loutfi]{unsupervised}
Martin L\"{a}ngkvist, Lars Karlsson, and Amy Loutfi.
\newblock Sleep stage classification using unsupervised feature learning.
\newblock \emph{Adv. Artif. Neu. Sys.}, 2012:\penalty0 5:5--5:5, January 2012.
\newblock ISSN 1687-7594.
\newblock \doi{10.1155/2012/107046}.
\newblock URL \url{http://dx.doi.org/10.1155/2012/107046}.

\bibitem[Li et~al.(2017)Li, Liu, Chen, and
  Rudin]{DBLP:journals/corr/abs-1710-04806}
Oscar Li, Hao Liu, Chaofan Chen, and Cynthia Rudin.
\newblock Deep learning for case-based reasoning through prototypes: {A} neural
  network that explains its predictions.
\newblock \emph{CoRR}, abs/1710.04806, 2017.
\newblock URL \url{http://arxiv.org/abs/1710.04806}.

\bibitem[Lipton(2016)]{DBLP:journals/corr/Lipton16a}
Zachary~Chase Lipton.
\newblock The mythos of model interpretability.
\newblock \emph{CoRR}, abs/1606.03490, 2016.
\newblock URL \url{http://arxiv.org/abs/1606.03490}.

\bibitem[Massimini et~al.(2004)Massimini, Huber, Ferrarelli, Hill, and
  Tononi]{sws_algo2}
Marcello Massimini, Reto Huber, Fabio Ferrarelli, Sean Hill, and Giulio Tononi.
\newblock The sleep slow oscillation as a traveling wave.
\newblock \emph{Journal of Neuroscience}, 24\penalty0 (31):\penalty0
  6862--6870, 2004.

\bibitem[Paszke et~al.(2017)Paszke, Gross, Chintala, Chanan, Yang, DeVito, Lin,
  Desmaison, Antiga, and Lerer]{paszke2017automatic}
Adam Paszke, Sam Gross, Soumith Chintala, Gregory Chanan, Edward Yang, Zachary
  DeVito, Zeming Lin, Alban Desmaison, Luca Antiga, and Adam Lerer.
\newblock Automatic differentiation in pytorch.
\newblock In \emph{NIPS-W}, 2017.

\bibitem[Pedregosa et~al.(2011)Pedregosa, Varoquaux, Gramfort, Michel, Thirion,
  Grisel, Blondel, Prettenhofer, Weiss, Dubourg, Vanderplas, Passos,
  Cournapeau, Brucher, Perrot, and Duchesnay]{scikit-learn}
F.~Pedregosa, G.~Varoquaux, A.~Gramfort, V.~Michel, B.~Thirion, O.~Grisel,
  M.~Blondel, P.~Prettenhofer, R.~Weiss, V.~Dubourg, J.~Vanderplas, A.~Passos,
  D.~Cournapeau, M.~Brucher, M.~Perrot, and E.~Duchesnay.
\newblock Scikit-learn: Machine learning in {P}ython.
\newblock \emph{Journal of Machine Learning Research}, 12:\penalty0 2825--2830,
  2011.

\bibitem[Priebe et~al.(2003)Priebe, Marchette, DeVinney, and
  Socolinsky]{priebe2003classification}
Carey~E Priebe, David~J Marchette, Jason~G DeVinney, and Diego~A Socolinsky.
\newblock Classification using class cover catch digraphs.
\newblock \emph{Journal of classification}, 20\penalty0 (1):\penalty0 003--023,
  2003.

\bibitem[Schutte-Rodin et~al.(2008)Schutte-Rodin, Broch, Buysse, Dorsey, and
  Sateia]{insomnia_percentage}
Sharon Schutte-Rodin, Lauren Broch, Daniel Buysse, Cynthia Dorsey, and Michael
  Sateia.
\newblock Clinical guideline for the evaluation and management of chronic
  insomnia in adults.
\newblock \emph{Journal of Clinical Sleep Medicine}, 4\penalty0 (05):\penalty0
  487--504, 2008.

\bibitem[Snell et~al.(2017)Snell, Swersky, and
  Zemel]{DBLP:journals/corr/SnellSZ17}
Jake Snell, Kevin Swersky, and Richard~S. Zemel.
\newblock Prototypical networks for few-shot learning.
\newblock \emph{CoRR}, abs/1703.05175, 2017.
\newblock URL \url{http://arxiv.org/abs/1703.05175}.

\bibitem[Sors et~al.(2018)Sors, Bonnet, Mirek, Vercueil, and Payen]{cnn}
Arnaud Sors, St{\'e}phane Bonnet, S{\'e}bastien Mirek, Laurent Vercueil, and
  Jean-Fran{\c{c}}ois Payen.
\newblock A convolutional neural network for sleep stage scoring from raw
  single-channel eeg.
\newblock \emph{Biomedical Signal Processing and Control}, 42:\penalty0
  107--114, 2018.

\bibitem[Stephansen et~al.(2018)Stephansen, Olesen, Olsen, Ambati, Leary,
  Moore, Carrillo, Lin, Han, Yan, et~al.]{nature}
Jens~B Stephansen, Alexander~N Olesen, Mads Olsen, Aditya Ambati, Eileen~B
  Leary, Hyatt~E Moore, Oscar Carrillo, Ling Lin, Fang Han, Han Yan, et~al.
\newblock Neural network analysis of sleep stages enables efficient diagnosis
  of narcolepsy.
\newblock \emph{Nature communications}, 9\penalty0 (1):\penalty0 5229, 2018.

\bibitem[Vallat(2019)]{yasa}
Raphael Vallat.
\newblock raphaelvallat/yasa: v0.1.3, March 2019.
\newblock URL \url{https://doi.org/10.5281/zenodo.2583431}.

\bibitem[Viera et~al.(2005)Viera, Garrett, et~al.]{kappa_meaning}
Anthony~J Viera, Joanne~M Garrett, et~al.
\newblock Understanding interobserver agreement: the kappa statistic.
\newblock \emph{Fam med}, 37\penalty0 (5):\penalty0 360--363, 2005.

\end{thebibliography}
\newpage
\appendix

\section*{Appendix A. Feature Selected from MGH Dataset} 

\begin{table}[htb]
    \centering
\resizebox{\textwidth}{!}{%
\begin{tabular}{l@{\quad}cccccccc}
  \toprule
  \multirow{2}{*}{\raisebox{-\heavyrulewidth}{Channels}} & \multicolumn{8}{c}{Features} \\
  \cmidrule{2-9}
  & Spindle & SWS & Delta & Theta & Alpha & Beta & Kurtosis & Amplitude \\
  \midrule
 F3-M2    &  &3  &4  &4  &4  &4  &  &4 \\
 F4-M1    &  &3  &4  &4  &4  &  &  &4\\
 C3-M2    &  &3  &4  &4  &  &  &  &4\\
 C4-M1    &  &3  &4  &  &  &  &  &4\\
 O1-M2    &  &2  &4  &4  &  &  &  &4\\
 O2-M1    &  &2  &4  &4  &  &  &  &4\\
 E1-M2    &  &  &  &  &4  &  &  &4\\
 E2-M2    &  &  &  &  &  &  &  &\\
 Chin1-Chin2    &  &  &  &  & &  &  &\\
  \bottomrule
\end{tabular}}
    \caption{96 selected rules out of 240 expert rules using MGH Dataset} 
    \label{tab:features_mgh}
\end{table} 

\end{document}